\documentclass{article}

    \usepackage[preprint]{aaj2026}

\usepackage[utf8]{inputenc} % allow utf-8 input
\usepackage[T1]{fontenc}    % use 8-bit T1 fonts
\usepackage{hyperref}       % hyperlinks
\usepackage{url}            % simple URL typesetting
\usepackage{booktabs}       % professional-quality tables
\usepackage{amsfonts}       % blackboard math symbols
\usepackage{nicefrac}       % compact symbols for 1/2, etc.
\usepackage{microtype}      % microtypography
\usepackage{xcolor}         % colors

\usepackage{graphicx}
\usepackage{subcaption}
\usepackage{amsmath}

\title{VocaDet: Sample-Driven Open-Vocabulary Object Detection and Segmentation via Visual Tokenization and Vector Database Retrieval}

\author{%
    ZhiXin Sun\\
  PowerChina Zhongnan Engineering Corporation Limited \\
  \texttt{sunzxjdi@gmail.com} 
}

\begin{document}

\maketitle

\begin{abstract}
Open-vocabulary object detection and segmentation aim to recognize arbitrary objects beyond predefined categories. Although recent vision-language and reference-based approaches have significantly advanced this field, they often rely on text prompts, limited visual examples, or expensive feature matching procedures, making them difficult to scale to large and continuously expanding object repositories. In this work, we propose VocaDet, a sample-driven open-vocabulary object detection and segmentation framework that learns object concepts directly from user-provided positive and negative sample collections without model retraining. The key idea is to transform continuous visual representations into discrete visual vocabularies and perform efficient retrieval-based recognition through a scalable vector database. Specifically, we employ DINOv3 as the visual feature extractor and apply agglomerative clustering with adaptive clustering sensitivity to generate multi-granularity visual tokens. These visual tokens, together with position-debiased representations and spatial topology information, are stored as expandable object memories in a vector database. During inference, query images are converted into visual tokens and efficiently matched against the stored object memories for object localization and segmentation. Furthermore, a background filtering mechanism is introduced to remove frequently occurring background patterns and reduce redundant retrieval operations in practical fixed-camera scenarios. Experiments on the UA-DETRAC dataset demonstrate that VocaDet achieves effective open-vocabulary detection performance without conventional detector training, while supporting continuously expandable recognition capability as additional positive and negative samples are accumulated. Code is available at \url{https://github.com/sunzx97/VocaDet}.
\end{abstract}

\section{Introduction}
Object detection and segmentation are among the most fundamental and widely applied tasks in computer vision. Owing to its strong real-time performance and mature engineering deployment, the YOLO series\cite{redmon2018yolov3, wang2024yolov10} has been extensively adopted in industrial applications. However, such methods typically require large-scale data collection and task-specific training for different scenarios, and inevitably suffer from false positives and false negatives.

Recently, methods such as SAM3\cite{carion2025sam}, Grounding DINO\cite{liu2024grounding}, T-Rex2\cite{jiang2403t}, DINO-X\cite{ren2024dino}, and INSID3\cite{cuttano2026insid3} have introduced context-aware detection and segmentation via text prompts or visual prompts, enabling open-vocabulary recognition. Nevertheless, these approaches usually require explicit visual prompt inputs or repeated similarity computation between visual prompts and target images, which limits their scalability to large reference contexts and prevents efficient inference.

In addition, recent approaches such as Rex-Thinker~\cite{jiang2025rex} and Rex-Omni~\cite{jiang2026detect} exploit multimodal large language models to further improve open-domain detection capabilities.
Most relevant to our approach is Training-free~\cite{espinosa2025no}, which introduces a training-free reference image-driven instance segmentation framework.
By leveraging frozen DINOv2\cite{oquab2023dinov2} semantic features and the category-agnostic segmentation capability of SAM2\cite{ravi2025sam}, it constructs a feature memory bank with feature aggregation and semantic-aware matching.
Given only a small number of reference images, this method achieves cross-domain generalization for automatic instance-level segmentation.

However, existing approaches remain limited in scenarios where users provide large-scale positive and negative sample collections and expect the system to automatically learn from these samples for efficient and accurate target retrieval in arbitrary images.

To address these limitations, we propose a large-scale sample-driven open-vocabulary object detection and segmentation framework. The method learns target concepts directly from user-provided large positive and negative sample repositories and performs efficient matching and detection in target images.

Specifically, we employ DINOv3\cite{simeoni2025dinov3} as the visual feature extractor and apply agglomerative clustering\cite{mullner2011modern} on extracted features to quantize images into discrete visual tokens. By applying different clustering thresholds, we further construct a hierarchical representation of visual tokens, forming both discrete token sets and their corresponding topological structures. The same pipeline is applied to positive and negative sample sets, converting them into visual token nodes or topological graphs, which are then stored in a vector database.

During inference, the input image or video is first transformed into a visual token representation. Each token is then retrieved against the vector database to find the most relevant positive and negative sample tokens. If a token node or local topological structure matches the stored representations with similarity above a predefined threshold, the corresponding region is identified as a target object.

The proposed method continuously improves as users incrementally expand the positive and negative sample repositories, leading to progressively better recognition performance. Meanwhile, by leveraging efficient vector database retrieval, the system enables fast detection and segmentation under large-scale sample settings.

\section{Method}
\begin{figure}[htbp]
    \centering
    \includegraphics[width=\linewidth]{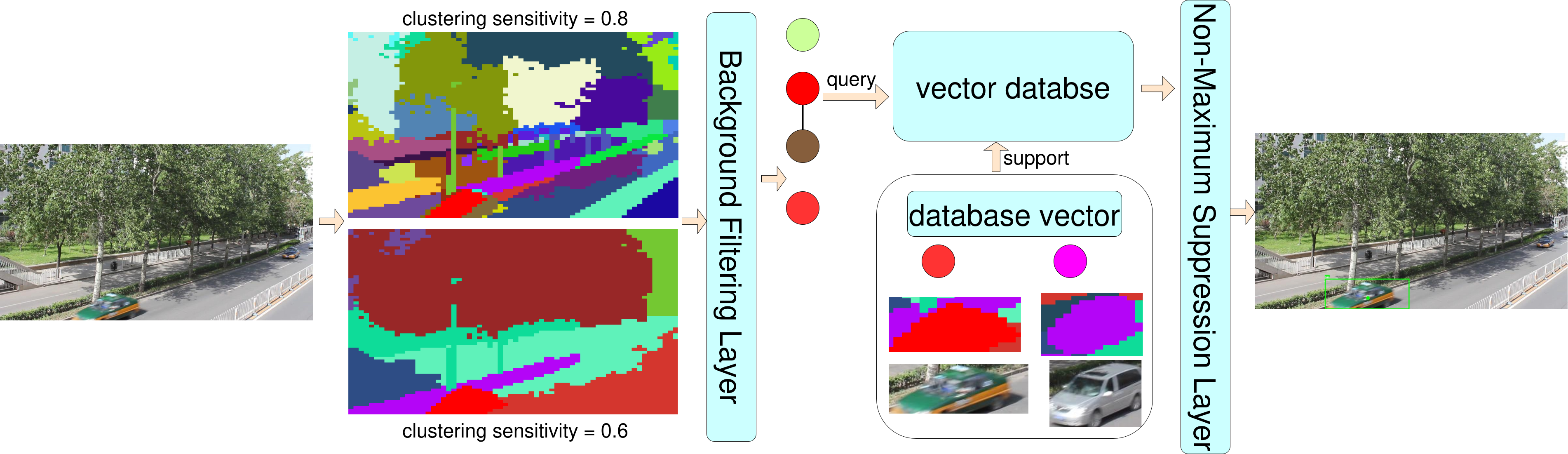}
    \caption{Overview of the proposed VocaDet framework. Given an input image, DINOv3 is first used to extract dense visual features, which are discretized into visual words through agglomerative clustering. A background filtering layer removes irrelevant visual words, while the remaining foreground visual words are retrieved against the vector database for object matching. The matched visual clusters are then converted into bounding boxes, followed by Non-Maximum Suppression (NMS) to obtain the final detection results.}
    \label{figframework}
\end{figure}

The overall architecture of the proposed VocaDet framework is illustrated in Fig.~\ref{figframework}.
Given an input image, DINOv3~\cite{simeoni2025dinov3} is first utilized to extract dense semantic visual features.
Unlike conventional object detectors that rely on predefined object categories, VocaDet constructs a data-driven visual vocabulary by discretizing continuous feature representations.
Specifically, we apply agglomerative clustering~\cite{mullner2011modern} to the extracted features and introduce a clustering sensitivity parameter to adaptively control the granularity of the generated visual words.
To improve detection efficiency, a background filtering layer is introduced to remove visual words associated with irrelevant regions.
The remaining foreground visual words are then used to perform similarity-based retrieval in a vector database such as Milvus\cite{wang2021milvus}, where both foreground and background object prototypes are stored.
By comparing the retrieved candidates, if a foreground object representation satisfies the predefined similarity threshold, the corresponding visual words are regarded as evidence of the target object.
The spatial clusters corresponding to these matched visual words are further exploited to localize objects and generate preliminary bounding boxes.
Finally, a Non-Maximum Suppression (NMS) layer is applied to merge highly overlapped predictions and obtain the final detection results.

We build the vector database based on INSID3~\cite{cuttano2026insid3}, which consists of three main stages: 
visual feature extraction, visual vocabulary construction, and position-debiased feature storage.

Specifically, given an input image, we first extract dense visual features using DINOv3~\cite{simeoni2025dinov3}.
The extracted features are then grouped into visual words through an agglomerative clustering algorithm~\cite{mullner2011modern}.
Since the same object may contain spatially disconnected regions with similar visual characteristics, 
we further separate disconnected clusters with identical visual patterns into independent object instances according to their spatial connectivity.

To reduce the influence of spatial bias during feature matching, we adopt the position debiasing strategy proposed in INSID3~\cite{cuttano2026insid3} to transform the original DINOv3 features into position-debiased representations.
Both the features stored in the vector database and the query features during inference are represented using these position-debiased features.

The construction of the vector database requires human annotation.
The database is designed to store both target objects (positive samples) and background regions (negative samples).
Given an input image, the visual features are first extracted by DINOv3~\cite{simeoni2025dinov3} and discretized into visual words using agglomerative clustering~\cite{mullner2011modern}.
Users can then specify target regions through different annotation forms, including points, bounding boxes, and masks.

For point-based annotation, the cluster containing the selected point is directly added to the vector database.
For bounding-box or mask-based annotation, only clusters whose regions are mostly contained within the annotated area are selected.
Specifically, if more than 90\% of a cluster region is covered by the bounding box or mask, the corresponding cluster feature is stored as an object representation.

When multiple clustered regions are included in the same annotation area, their spatial topology is additionally preserved in the vector database.
Specifically, if two clusters satisfy the coverage constraint and are spatially connected, their mutual connectivity relationship is recorded in an \textit{indexList} field.
For example, if Cluster 1 and Cluster 2 are both contained within the annotated object region and Cluster 1 is connected to Cluster 2, the feature entries of both clusters are stored in the database.
Meanwhile, the identifier of Cluster 2 is added to the \textit{indexList} of Cluster 1, and the identifier of Cluster 1 is added to the \textit{indexList} of Cluster 2.

During inference, a query visual word is considered a valid match only when both feature similarity and topological consistency are satisfied.
For example, suppose two query clusters $A$ and $B$ achieve similarity scores higher than the predefined threshold with Cluster 1 and Cluster 2 stored in the vector database, respectively.
The retrieval result is accepted only if Cluster $A$ and Cluster $B$ preserve the same adjacency relationship as Cluster 1 and Cluster 2.
In this case, the matched clusters $A$ and $B$ are jointly regarded as a detected object instance.

Since different clustering sensitivities produce different visual vocabulary granularities, 
we construct multiple vector databases corresponding to different clustering sensitivity settings.
During inference, the query features are clustered using the same clustering sensitivity as the target database, 
and retrieval is performed only within the corresponding vector database.
This strategy enables flexible object representation and retrieval under different granularity levels.

In practical engineering scenarios, surveillance cameras are often deployed at fixed locations, where the captured scenes mainly consist of static background regions for most of the time.
Directly querying all visual words against the vector database in such scenarios would introduce substantial redundant computation, especially when a large number of background features are repeatedly retrieved.
To improve detection efficiency, we introduce a background filtering layer to remove background-related visual words before vector database retrieval.

Specifically, the background filtering layer maintains a background feature memory that stores representative background visual words.
Given an input image, DINOv3~\cite{simeoni2025dinov3} is first used to extract visual features, which are then discretized into visual words through an agglomerative clustering algorithm~\cite{mullner2011modern}.
The generated visual words are initially compared with all background vectors stored in the background feature memory.
If the similarity score between a query visual word and a background vector exceeds a predefined threshold, the visual word is regarded as a background region and excluded from subsequent vector database retrieval.
Only the remaining visual words are forwarded to the object vector database for further matching.

To adapt to dynamic environmental changes, the background feature memory is updated online.
For each input frame, the extracted visual words are compared with all vectors stored in the current background feature memory.

The update process follows a matching-and-replacement strategy.
Specifically, background vectors whose similarity scores with the current frame exceed the predefined threshold are considered matched and retained.
The unmatched background vectors are removed from the memory, while the visual words from the current frame that successfully match the background memory are regarded as background observations.

The remaining unmatched visual words are further processed by the object vector database.
If an unmatched visual word is retrieved as a negative sample (background vector) stored in the object vector database, or achieves a similarity score higher than the predefined threshold with the most relevant positive object vector, it is considered a background feature and added to the background feature memory.

Through this online background filtering mechanism, redundant background queries are effectively eliminated before vector database retrieval, significantly reducing computational overhead while maintaining accurate object detection performance.

The matched visual clusters retrieved from the vector database are first converted into object bounding boxes according to their spatial distributions.
However, since VocaDet performs clustering under multiple clustering sensitivity settings, the same image may generate visual words with different granularities.
These multi-granularity visual words are independently matched with their corresponding vector databases, which may lead to multiple overlapping bounding boxes for the same object.

To obtain consistent detection results, a Non-Maximum Suppression (NMS) layer is applied after vector retrieval.
By considering the confidence scores and spatial overlap between candidate boxes, redundant detections are suppressed, and the remaining bounding boxes are selected as the final object predictions.

\section{Experiments}
We evaluate the proposed VocaDet framework on the UA-DETRAC dataset~\cite{wen2020ua}.
The experimental results demonstrate that VocaDet achieves competitive detection performance without requiring conventional model training.
More importantly, unlike fixed-category detectors, the performance of VocaDet can be continuously improved by expanding the positive and negative sample sets stored in the vector database.
As more representative object examples are accumulated, the constructed visual memory becomes more comprehensive, leading to more accurate target retrieval.

However, we also observe an inherent limitation of the proposed approach.
Specifically, when multiple spatially adjacent objects belonging to the same category appear in an image, they may be incorrectly merged into a single detection instance.
This issue occurs because the proposed method primarily relies on visual feature similarity and spatial clustering, which lack sufficient discriminative information to separate neighboring instances with highly similar appearances.

\section{Conclusion}
In this work, we propose VocaDet, a sample-driven open-vocabulary object detection and segmentation framework that enables flexible target recognition through visual tokenization and vector database retrieval. By leveraging DINOv3 visual representations, adaptive agglomerative clustering, position-debiased feature representations, and topology-aware object memories, VocaDet learns object concepts directly from user-provided positive and negative samples without requiring model retraining. Furthermore, the proposed background filtering mechanism improves retrieval efficiency in practical fixed-camera scenarios by reducing redundant background queries. Experimental results on the UA-DETRAC dataset demonstrate that VocaDet provides an effective and scalable solution for open-vocabulary perception, where recognition capability can be continuously improved by expanding the sample repository. However, as a sample-driven framework, the initial construction of the vector database may suffer from the cold-start problem, where limited samples make it difficult to accurately characterize the distribution of object categories and establish reliable feature boundaries. In future work, we will investigate the distribution patterns of samples within the vector database and explore adaptive strategies for organizing visual memories and discovering more discriminative category boundaries, thereby reducing the dependency on manually collected samples during the initial deployment stage.
\medskip

\bibliography{biblio.bib}

\end{document}